\documentclass[runningheads]{llncs}
\usepackage[T1]{fontenc}
\usepackage{graphicx}
\usepackage{tikz}
\usepackage{pgf-pie} 
\usepackage{xcolor}
\usepackage{amsmath}
\usetikzlibrary{decorations.text}
\usetikzlibrary{calc}
\usetikzlibrary{fit}
\usetikzlibrary{bayesnet}
\usetikzlibrary{arrows}
\usetikzlibrary {arrows.meta}
\usetikzlibrary{positioning}
\usetikzlibrary{shapes}
\usetikzlibrary{shadows}
\usepackage{adjustbox}
\usepackage{multirow}
\usepackage{makecell}
\usepackage{pgfplotstable} 
\pgfplotsset{compat=1.17}
\usepackage{subcaption}
\usepackage{booktabs}
\usetikzlibrary{arrows}
\usepackage[hidelinks]{hyperref}

\begin{document}

\title{Predicting the Road Ahead: A Knowledge Graph based Foundation Model for Scene Understanding in Autonomous Driving}

% \subtitle{Leveraing Pre-trained Language Models for Scene Understanding in Autonomous Driving}

% \subtitle{A Foundation Model for Scene Understanding in Autonomous Driving based on Knowledge Graphs}

% \subtitle{(Toward) A Knowledge-graph-based Foundation Model for Scene Understanding in Autonomous Driving}

% \subtitle{Predicting the Road Ahead: A Knowledge-graph-based Foundation Model for Scene Understanding in Autonomous Driving}

%
\titlerunning{A Knowledge Graph based Foundation Model for Scene Understanding}
% \author{
% Hongkuan Zhou\inst{1,2}\orcidID{0000-0002-3665-9822} \and
% Stefan Schimid\inst{1} \and 
% Yicong Li\inst{3} \and \\
% Lavdim Halilaj\inst{1} \and
% Xiangtong Yao\inst{4}\orcidID{0000-0003-2556-3072} \and
% Wei Cao\inst{1,2}\orcidID{0009-0005-5163-6484}
% }
\author{
Hongkuan Zhou\inst{1,2}\and
Stefan Schimid\inst{1} \and 
Yicong Li\inst{3} \and
Lavdim Halilaj\inst{1} \and \\
Xiangtong Yao\inst{4} \and
Wei Cao\inst{1,2}
}

\authorrunning{H. Zhou et al.}

\institute{Bosch Corporate Research, Renningen, Germany 
\email{\{hongkuan.zhou,stefan.schmid5,lavdim.halilaj,wei.cao4\}@de.bosch.com} \and
University of Stuttgart, Stuttgart, Germany \and
University of Montreal, Montreal, Canada\\
\email{yi.cong.li@umontreal.ca} \and
Technical University of Munich, Munich, Germany\\
\email{xiangtong.yao@tum.de }
}

\maketitle              % typeset the header of the contribution
\begin{abstract}
\iffalse
This paper proposes a novel methodology for training a foundation model (FM) for scene understanding in autonomous driving. The approach leverages large language models (LLMs) and uses a symbolic scene representation based on knowledge graphs (KGs). This approach facilitates the integration of diverse autonomous driving datasets and the training of large FMs by transforming driving scenes into a uniform scene representation.
The model uses a bird's eye view (BEV) symbolic representation extracted from the KG for each driving scene. The BEV representation is serialized into a sequence of tokens capturing entity names, spatial layout, and environmental conditions.
The paper investigates the capability of LLMs to understand driving scenes and generate predictions on the next scene. 
Experiments were conducted using the nuScenes dataset and KG. The results demonstrate that fine-tuned models achieve significantly higher accuracy in all tasks. The fine-tuned T5 model achieved a next scene prediction accuracy of $86.7\%$. The ablation studies highlight the importance of additional metadata, such as traveled distance and orientation difference, for accurate next scene prediction. This paper concludes that LLMs offer a promising foundation for developing more powerful models for scene understanding in autonomous driving.
\fi

The autonomous driving field has seen remarkable advancements in various topics, such as object recognition, trajectory prediction, and motion planning. However, current approaches face limitations in effectively comprehending the complex evolutions of driving scenes over time. This paper proposes FM4SU, a novel methodology for training a symbolic foundation model (FM) for scene understanding in autonomous driving. It leverages knowledge graphs (KGs) to capture sensory observation along with domain knowledge such as road topology, traffic rules, or complex interactions between traffic participants. A bird’s eye view (BEV) symbolic representation is extracted from the KG for each driving scene, including the spatio-temporal information among the objects
across the scenes. The BEV representation is serialized into a sequence of tokens and given to pre-trained language models (PLMs) for learning an inherent understanding of the co-occurrence among driving scene elements and generating predictions on the next scenes. We conducted a number of experiments using the nuScenes dataset and KG in various scenarios. The results demonstrate that fine-tuned models achieve significantly higher accuracy in all tasks. The fine-tuned T5 model achieved a next scene prediction accuracy of  86.7\%. 
This paper concludes that FM4SU offers a promising foundation for developing more comprehensive models for scene understanding in autonomous driving.

\end{abstract}
\definecolor{network-blue}{RGB}{165, 192, 221}
\definecolor{light-yellow}{RGB}{238, 233, 218}
\definecolor{light-green}{RGB}{129, 184, 113}
\definecolor{light-red}{RGB}{242, 182, 160}
\definecolor{light-blue}{RGB}{124, 150, 171}
\definecolor{light-orange}{RGB}{255,147,0}
\definecolor{light-purple}{RGB}{150,115,166}
\definecolor{dark-green}{RGB}{85, 124, 86}
\definecolor{dark-red}{RGB}{217, 22, 86}
\definecolor{purple}{RGB}{155, 126, 189}
\definecolor{dark-purple}{RGB}{59, 30, 84}
\section{Introduction}
\begin{figure*}[tb]
\centering
\includegraphics[width=12cm]{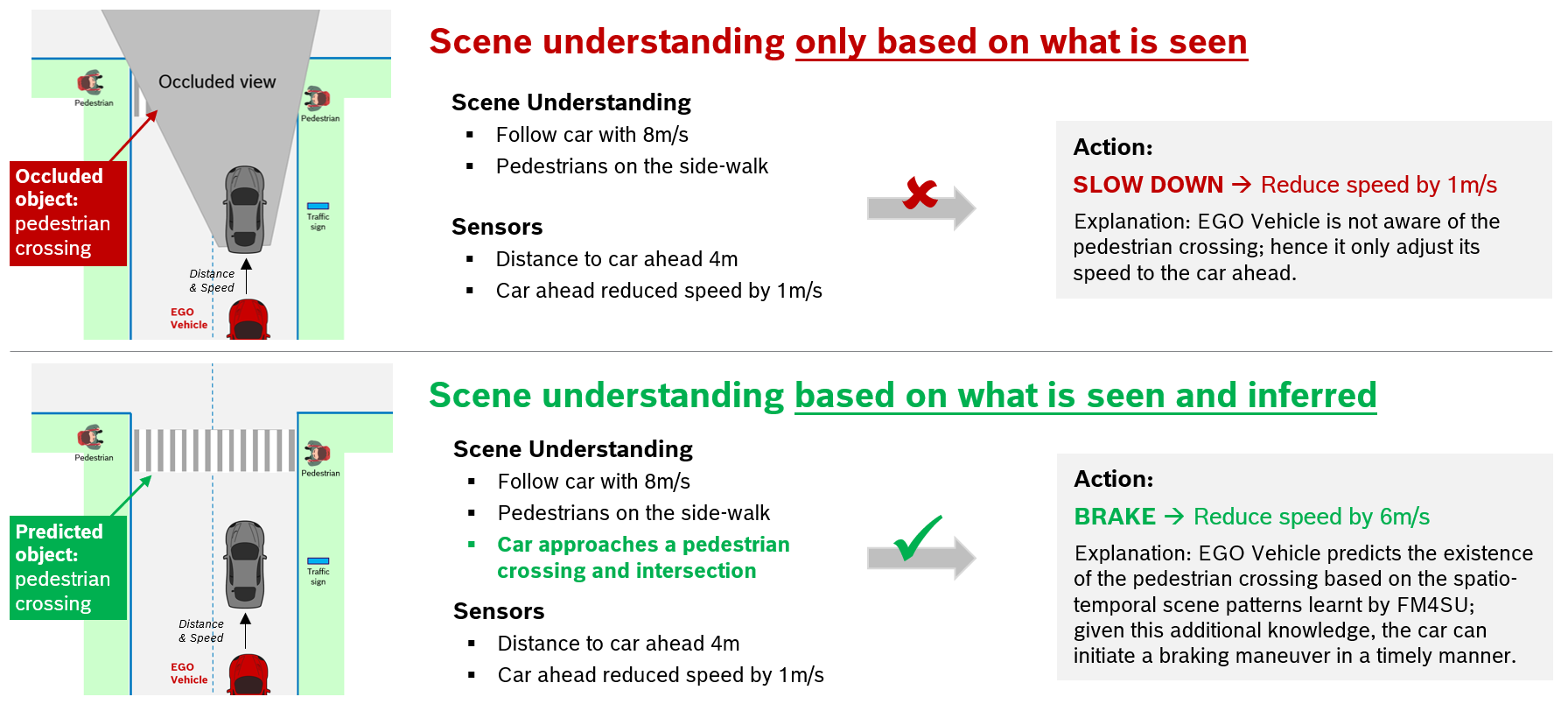}
\caption{\small{\textbf{Motivation} -- An exemplary driving scene illustrating how FM4SU is able to enrich scene understanding. 
By learning spatio-temporal patterns within and across diverse driving scenarios, FM4SU can infer missing information (e.g. pedestrian crossing). This enables more informed and consequently safer decision-making for the vehicle.}}
\label{fig:motivation}
\end{figure*}

Multi-modal foundation models are receiving increasing attention in autonomous driving for their ability to perform tasks such as object detection, prediction of future trajectories, and scene understanding~\cite{DBLP:journals/corr/abs-2402-01105}. Scene understanding in the context of autonomous driving refers to the ability of an autonomous car to perceive its surroundings, interpret the perceived information, and make informed decisions based on that interpretation~\cite{doi:10.1177/0278364913491297}.
Figure~\ref{fig:motivation} exemplifies the potential of employing scene understanding derived from sensory data and coupled with a specifically trained foundation models (FMs) to enhance safety in autonomous driving. The inclusion of additional contextual information, even if it's merely predicted by the FMs, is crucial for an autonomous vehicle to act appropriately.

Learning an exhaustive and easily reusable FMs for scene understanding in autonomous driving poses significant challenges. These complexities stem mainly from the diversity of sensor technologies, such as video and radar, varying ranges and resolutions, as well as sensor deployments (e.g., mounting positions) on vehicles. 
Consequently, the development and evaluation of a truly large FM from diverse datasets (e.g., nuScenes~\cite{DBLP:conf/cvpr/CaesarBLVLXKPBB20}) becomes a very challenging task.
To address these shortcomings, we advocate a novel ontology-based symbolic scene representation as input for learning the foundation model. 
This supports the integration of heterogeneous autonomous driving datasets into a uniform and compact driving scene representation~\cite{DBLP:journals/ijsc/HalilajLMHS23}, and thus, facilitates the training of a truly large FM. 
The advantages of using a symbolic representation are manifold. 
First, it provides a rich structure and semantics,  well suited to incorporate
additional domain knowledge via explicit relationships.
Second, it allows experts to easily validate the model and use it for the explainability of predictions. 
Third, it enables further predicting and validation of the results based on the axioms defined in the ontology. 
Finally, such a high-level scene representation aids in integrating the model-based predictions into diverse downstream tasks (e.g., object detection, semantic segmentation, trajectory prediction)~\cite{DBLP:journals/ijsc/HalilajLMHS23}.

We evaluate our approach on the basis of a real-world autonomous driving dataset, nuScenes~\cite{DBLP:conf/cvpr/CaesarBLVLXKPBB20}, and investigate to what extent the trained FM is able to predict masked entities in the scene and a complete next scene. 
Figure~\ref{fig:learning-pipeline} illustrates our learning pipeline. 
In the first step, we transform the driving scenes (all relevant information extracted by standard object detection and semantic segmentation algorithms) into a knowledge graph (KG). 
The KG captures the spatio-temporal relations in and across the scenes and the detected entities (both static and dynamic). Based on these semantically enriched scene representations, we extract for each scene a symbolic representation from a Bird's Eye View (BEV) around the Ego Vehicle (EV). These ontology-based scene representations are then used for the training of a large transformer model. Inspired by the recent breakthroughs in large language models (LLMs), we use attention mechanisms to train a large model to understand realistic driving scenes and their spatio-temporal evolution enfolding in the subsequent scene. 

We evaluate to what extent large transformer models are able to learn realistic driving scenes and generate predictions on the next scene. 
Further, we investigate different approaches, including fine-tuning existing large models and different model sizes. The main \emph{contributions} of our work are shown as follows: 
%and ... TODO
\input{tikz/pipeline}
\begin{itemize}
\item A new methodology for learning a foundation model for scene understanding based on an ontology-based symbolic scene representation. The proposed approach enables easy integration of diverse datasets (different sources, with different sensor modalities, ...) and facilitates the training large FMs. 
\item We conducted a number of experiments to evaluate to what extent large (language) models are capable of understanding driving scenes without additional visual modalities, such as RGB images or LiDAR, by predicting masked areas in a given scene and/or predicting the next scenes.
\item A new dataset and tools\footnote{See here: \url{https://github.com/boschresearch/fm4su}} for the research community to develop and benchmark new algorithms for learning a foundation model for scene understanding based on the nuScenes dataset.
\end{itemize}

% \section{Preliminary}

% \subsection{Problem Statement -- TODO}

% A matrix of cells with traffic objects/entities is extracted from the driving
% scenes KG. (b) The matrices extracted from the scenes at time steps T and T + 1 are
% converted into serialized sequences of tokens. (c) The Transformer model is trained
% using either masked prediction of (illustrated on the left) or next scene prediction
% (illustrated on the right

\section{Related Work}
The use of LLMs for addressing tasks representable through language has garnered increasing attention in diverse fields, including autonomous driving~\cite{2,DBLP:journals/ral/XuZXZGWLZ24,10297415,Cui_2024_WACV}, protein 3D structure representation~\cite{3}, and robot manipulation~\cite{1,zhou2023language}. In the domain of autonomous driving, LLMs have been employed to tackle challenges such as motion prediction~\cite{DBLP:conf/eccv/ZhouHBZLQZGQL24,DBLP:journals/corr/abs-2309-16534}, traffic visual question answering~\cite{DBLP:conf/eccv/ZhouHBZLQZGQL24,DBLP:journals/corr/abs-2407-00959}, and end-to-end autonomous driving~\cite{DBLP:journals/ral/XuZXZGWLZ24,DBLP:journals/corr/abs-2407-00959,DBLP:conf/itsc/ZhouSSL23}. Additionally, knowledge graphs (KGs) have been explored as a complementary approach, providing the relationships between entities in traffic scenarios~\cite{DBLP:journals/ijsc/HalilajLMHS23,DBLP:journals/ral/SunWHL24}. The subsequent subsections provide a detailed discussion of scene understanding, FMs, and KGs in the context of autonomous driving.

\subsection{Pre-trained Language Models}
PLMs are an early attempt to extract semantic meaning from natural language. ELMo~\cite{peters-etal-2018-deep} aims to capture context-aware word representations by first pre-training a bidirectional LSTM (biLSTM) network and then fine-tuning the biLSTM network according to specific downstream tasks. Moreover, drawing inspiration from the Transformer architecture \cite{vaswani2017attention} and incorporating self-attention mechanisms, BERT \cite{devlin2018bert} takes language model pre-training a step further. It accomplishes this by conducting bidirectional pre-training exercises on extensive unlabeled text corpora. These specially crafted pre-training tasks imbue BERT with contextual understanding, resulting in highly potent word representations. \\[-10pt]

% \subsection{Large Language Models}
LLMs become popular as the scaling of PLMs leads to improved performance on downstream tasks. 
Many researchers study the performance limits of PLMs by scaling the size of models and datasets~\cite{hoffmann2022training}, e.g., the comparatively small 1.5B parameter GPT-2~\cite{radford2019language} 
versus larger 175B GPT-3~\cite{brown2020language}, 450B Llama~\cite{llama} and 540B PaLM~\cite{chowdhery2022palm}. Although these models share a similar structure, the enlarged models display different behaviors and show surprising abilities compared to previous works.
Considering our available computational resources and the level of open-source accessibility, we opt for the Text-to-Text Transfer Transformer~\cite{DBLP:journals/jmlr/RaffelSRLNMZLL20} (T5) as a pre-trained language model to build a KG-based FM for scene understanding.

\subsection{Scene Understanding in Autonomous Driving}
Scene understanding is a critical aspect of autonomous driving, enabling vehicles to perceive and interpret their surroundings for safe and efficient navigation. Recent advancements in autonomous driving models~\cite{chitta2022transfuserimitationtransformerbasedsensor,zhou2023matters} leverage multi-modal inputs, including RGB images, LiDAR, and BEV representations. 
RGB images provide detailed visual context, LiDAR offers precise depth information, and BEV representations integrate these inputs into a structured spatial view, facilitating better scene interpretation. 
% Despite these advancements, challenges persist in comprehending complex and dynamic scenarios, such as ambiguous road layouts, dense traffic interactions, and environmental uncertainties. 
%
LLMs have emerged as promising tools to address these challenges by incorporating contextual predicting and improving interpretability. For instance, models like PlanAgent \cite{zheng2024planagent} employ hierarchical predicting with BEV inputs to generate interpretable motion commands, while DriveMLM \cite{wang2023drivemlm} integrates multi-view image data and LiDAR point clouds to produce high-level planning decisions. However, due to the inherent complexity of driving scenarios, LLMs sometimes struggle to fully understand intricate environments. Additionally, these approaches face limitations such as slow prediction speeds caused by iterative processes \cite{wen2023dilu} and the lack of diverse, realistic datasets for robust evaluation \cite{ding2023hilm}.
%Addressing these challenges is crucial to advance the scene understanding in autonomous driving.

\subsection{Foundation Models in Autonomous Driving}
With the development of LLMs and FMs, more research attention is being paid to integrating the common sense knowledge existing in FMs in the field of autonomous driving. Tian et al.~\cite{DBLP:journals/corr/abs-2407-00959} tokenizes multi-view videos, HD-maps, and symbolic representations of objects, enabling better utilization of LLM’s generalization capabilities to enhance autonomous vehicle planning in long-tail scenarios. Zhou et al.~\cite{DBLP:conf/eccv/ZhouHBZLQZGQL24} propose an Embodied Language Model (ELM), a comprehensive framework tailored for agents’ understanding of driving scenes with large spatial and temporal spans. DriveGPT4~\cite{DBLP:journals/ral/XuZXZGWLZ24} processes multi-frame video inputs alongside textual queries to predict low-level vehicle control signals and generate textual explanations in response. We observe a growing research trend in tokenizing multi-modal inputs—such as video, LiDAR data, and textual queries—and employing visual question answering in traffic scenes. In such a way, the knowledge embedded in pre-trained FMs is utilized to enhance the generalization capability of autonomous driving pipelines. In this work, we aim to investigate the spatial and temporal scene understanding ability of LLMs in autonomous driving, excluding additional visual modalities such as videos and LiDAR. This approach enables a clearer assessment of the benefits of incorporating LLMs’ knowledge in traffic scene analysis.

\subsection{Knowledge Graphs in Autonomous Driving}
KGs are getting more traction on their potential usage for different tasks in autonomous driving~\cite{DBLP:conf/kgswc/LuettinMHH22,zhou2025robustvisualrepresentationlearning}. 
The nuSceneKG~\cite{Mlodzian_2023_ICCV} is a KG that models all scene participants and road elements explicitly, as well as their semantic and spatial relationships. Based on that, SemanticFormer~\cite{DBLP:journals/ral/SunWHL24}  predicts multimodal
trajectories by predicting over a semantic traffic scene graph using a hybrid approach. SocialFormer~\cite{DBLP:journals/corr/abs-2405-03809} performs an agent interaction-aware trajectory prediction method that leverages the semantic relationship between the target vehicle and surrounding vehicles by making use of the road topology. 
In this paper, we build our symbolic BEV representation dataset for scene understanding from the nuSceneKG.

% \textit{
% Moved this here:
% In the field of Deep Learning, Transformer models have gained significant recognition recently thanks to the breakthroughs in large language models (LLMs) based on attention mechanism, such as GPT, BERT, etc. The idea of leveraging the capacity of LLMs to tackle other language-representable tasks has attracted growing attention, such as Autonomous driving~\cite{2}, Protein 3D structure representation~\cite{3}, and Robotic manipulation~\cite{1,zhou2023language}. For autonomous driving, the current stage is mostly focusing on computer vision (CV) field for tasks like object detection, semantic segmentation, and motion planning (\cite{4}). The models trained specifically for these tasks rely on camera, lidar, and radar data for low-level information extraction but often struggle to develop a full contextual understanding of the surrounding scene. Therefore, we want to mitigate this problem by leveraging the knowledge from a scene KG to finetune an LLM to achieve scene understanding, which  could bring potential for Level 4 autonomy.}
\section{Methodology}

\subsection{Knowledge Graph Representation}
\begin{figure*}[tb]
\centering
\includegraphics[width=\linewidth]{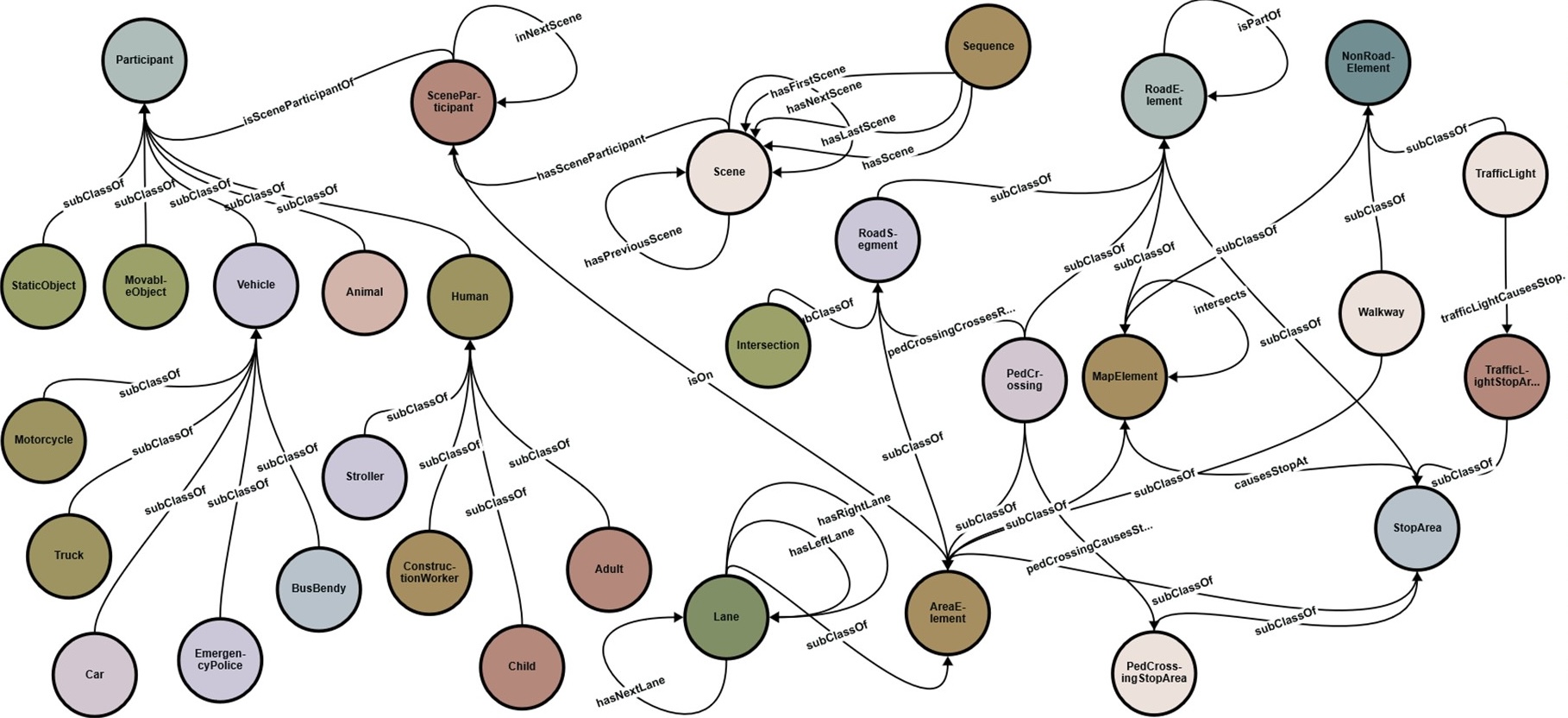}
\caption{\small \textbf{nuScenes Knowledge Graph} - a comprehensive representation of sensory data using ontological concepts with a strong focus on scene understanding. Scene objects $O$ are categorized based on different levels of abstraction and characteristics, such as static and dynamic for fixed and moving objects, respectively.}
\label{fig:nuScenesKG}
\end{figure*}

We leverage Knowledge Graphs as a medium to capture, organize and interlink the knowledge from the domain. 
It represents the information in form of triples $G = (E, R, E|V)$ where each element $e \in E$ is an entity, each $v \in V$ is a literal, e.g. number, date or text, and each $r \in R$ is a relation between two entities $<e_1 , r , e_2 >$ or an entity with a value $<e , r , v >$.

In our approach, we use the nuScenes knowledge graph (nuScenesKG\footnote{The ontology and the KG are available at \url{https://zenodo.org/records/10074393}.})~\cite{Mlodzian_2023_ICCV} to represent all information of the nuScenes dataset~\cite{DBLP:conf/cvpr/CaesarBLVLXKPBB20}.
The nuScenes dataset is one of the large-scale datasets for autonomous driving, comprising 1,000 driving sequences, each 20 seconds long, from Boston and Singapore. 
% It provides 360° coverage by the entire sensor suite of an autonomous vehicle consisting of 6 cameras, 1 LiDAR, 5 RADAR, GPS/IMU, and CAN bus data (e.g., velocity, acceleration, torque, steering angle, wheel speed). 
%It also contains accurate human-annotated maps details of the recorded areas. 
% The nuScenes ontology is built using human-annotated data, including map details.
% 
% Abstract concepts are defined by classes like $Sequence$, $Scene$, and $Participant$. 
% More specific classes describe occurring entities like $Car$, $Adult$, $Barrier$ or $Animal$, and map topology such as $Lane$, $Intersection$ and $StopArea$. 

The Ontology-(TBox) is structured in two main modules: 1-Agent: encompassing taxonomies of traffic participants: Vehicle (e.g., Car, Bus), Human (e.g., Adult, Child), relations capturing their state and location; and 2-Map: Road topology including its Segments, Lanes, Lane\_Snippets and Lane\_Slices. The expressivity level is SROIQ(D).
Overall, it comprises 42 classes, 10 object properties, and 24 datatype properties. 
Among them, \emph{Scene} is a key class, denoted as tuple $S_T = (B, C, T, P)$, where B - denotes its state, C - the relations to the predecessor and successor scene, respectively, T - denotes the specific time point at which the scene is captured and P - its participants including their geo-spatial relations.
Figure~\ref{fig:nuScenesKG} shows a small excerpt of the ontology depicting the core concepts and their relations.

Conversely, the KG-(ABox) persists instances and their respective states over time. Ontological axioms (transitivity, reflexivity and equivalence) enable the reasoning process to make many relations explicit, including the temporal evolution of scenes, participants, and scenery information. We perform additional operations to establish hierarchical relations between Lanes, Lane\_Snippets and Lane\_Slices. Further, we compute connections between agents (lateral, longitudinal) and their geo-spatial projection to the map topology at specific time points. 
%The nuScenes ontology along with the tools to construct the nuScenesKG from the dataset are available here XXX.
The KG comprises over 43 million triples in total, capturing the states and interlinks of static and dynamic entities over time.

\subsection{BEV Symbolic Scene Representation}
\input{tikz/architecture}
\begin{figure}[tb!]
    \centering
    \begin{subfigure}[b]{0.45\textwidth}
        \centering
        \includegraphics[width=\textwidth]{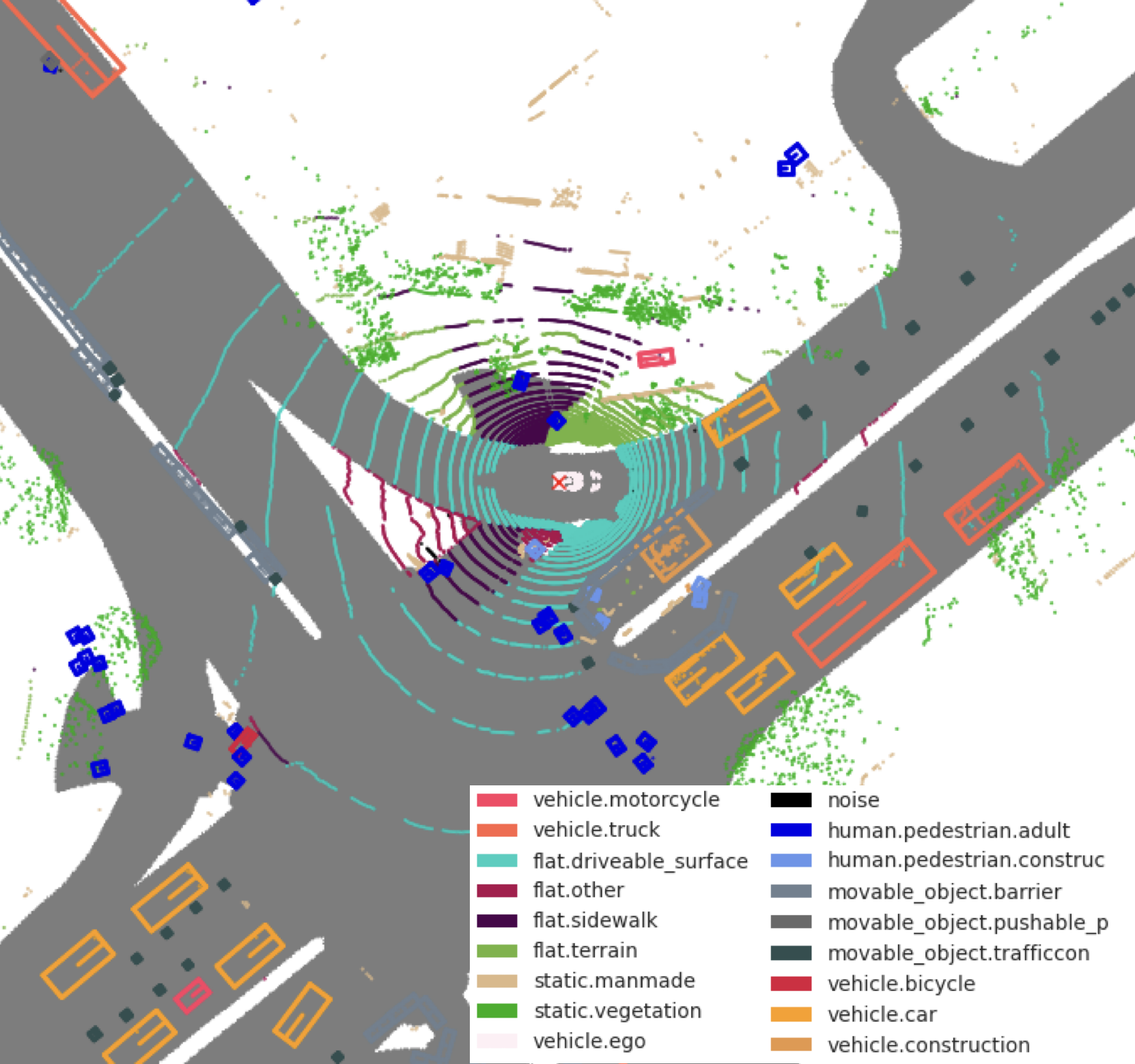}
        \caption{A Driving Scene}
        \label{fig:driving scene}
    \end{subfigure}
    % \hfill
    % \begin{subfigure}[b]{0.45\textwidth}
    %     \centering
    %     \includegraphics[width=\textwidth]{images/scene_vis_1_poly_w:o_b.png}
    %     \caption{Ego Vehicle's Perspective}
    %     \label{fig:ego vehicle}
    % \end{subfigure}
    \hfill
    \begin{subfigure}[b]{0.45\textwidth}
        \centering
        \includegraphics[width=\textwidth]{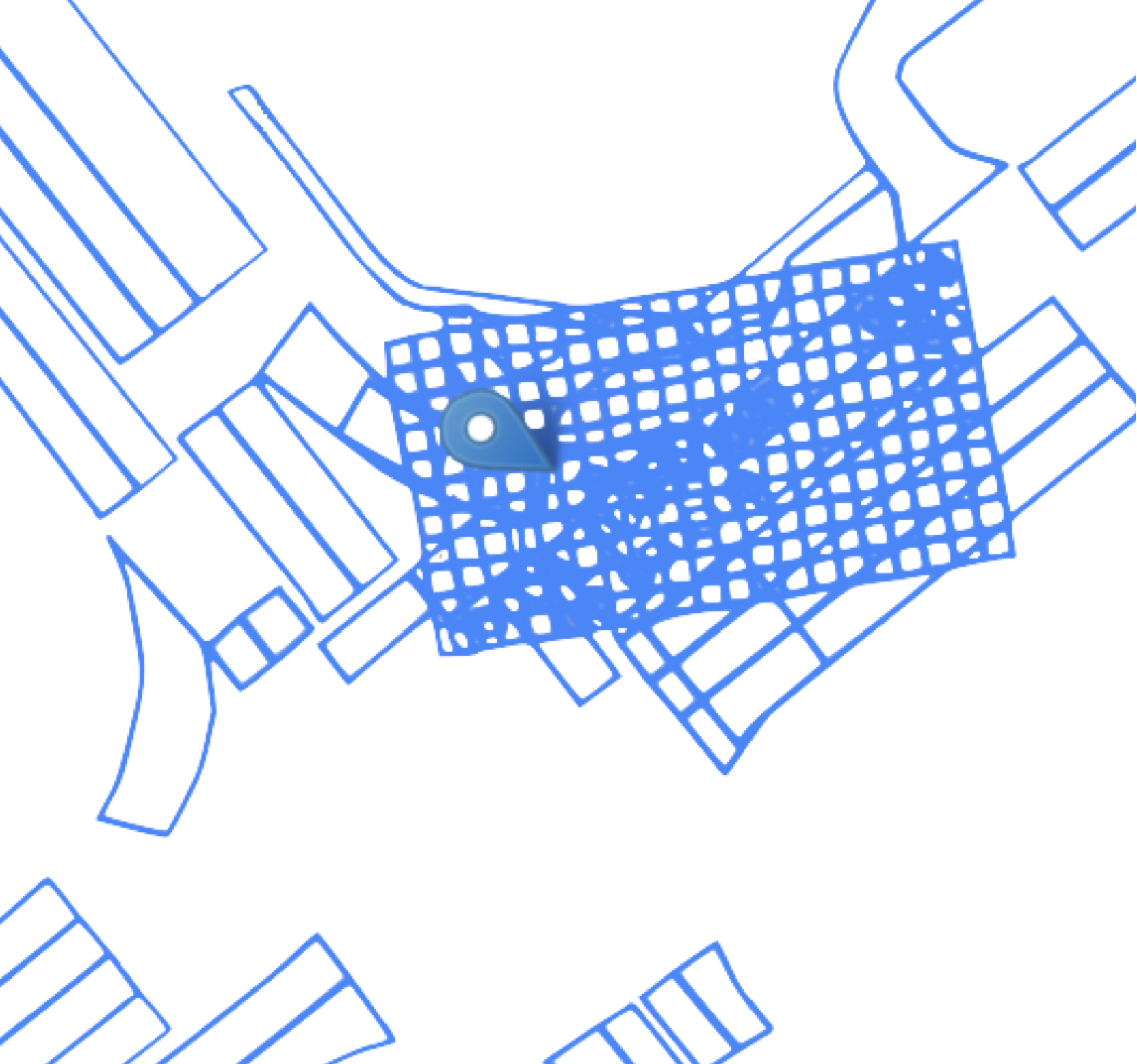}
        \caption{Matrix Representation}
        \label{fig:matrix}
    \end{subfigure}
    %  \hfill
    % \begin{subfigure}[b]{0.45\textwidth}
    %     \centering
    %     \includegraphics[width=\textwidth]{images/scene_vis_1_all_wo_m.png}
    %     \caption{Dynamic Entities}
    %     \label{fig:dynamic entities}
    % \end{subfigure}
    \vspace{-0.05cm}
    \caption{
        \small (a) A nuScenes visualization of a traffic scene using top-down LiDAR data, representing different entities with distinct colors. 
        % (b) A BEV representation of map concepts, including lanes, sidewalks, etc., is depicted as polygons. The pinpoint indicates the EV’s location. 
        (b) A BEV representation of map concepts, including lanes, sidewalks, etc., is depicted as polygons. A $20 \times 11$ area matrix $A$ to illustrate the scene range in our setup. The pinpoint indicates the EV’s location.
        % (d) Expanded with additional pinpoints for moving objects; each pinpoint represents a separate object.
    }. 
    \vspace{-2em}
    \label{fig:scene_comparison}
\end{figure}

The foundation model training is performed using a BEV-based scene representation extracted from the KG representation. 
As shown in Figure~\ref{fig:architecture}a, the geo-spatial area around the EV is represented as an area matrix $A_T$ for each driving Scene $S_T$. 
% We recommend a rectangular shape for the area matrices where the shorter side covers the area left and right of the EV, whereas the longer sides represents the back and front of the EV. 
The area matrix can be denoted as follows:
\begin{equation}
A_T = 
\begin{bmatrix}  
c^{T}_{n1} & c^{T}_{n2} & ... & c^{T}_{nm} \\
... & ... & ... & ... \\
c^{T}_{21} & c^{T}_{22} & ... & c^{T}_{2m} \\  
c^{T}_{11} & c^{T}_{12} & ... & c^{T}_{1m}  
\end{bmatrix}\text{,}
\end{equation}
where the dimensions of $A_T$ is $n \times m$ and 
% where $i = 1...n$ (for rows) and $j = 1...m$ (for columns).
each matrix cell $c^T_{ij}, \forall i \in \{1, 2, \dots, n\}, \\ \forall j \in \{1, 2, \dots, m\}$ represent a geographical area of height $h$ and width $w$. 
The height and width of each cell define the resolution of the matrix, whereas the number of rows and columns define the overall size or area of the scene. The exact area of cell $c^{T}_{ij}$ is defined by the coordinates $(a^{ij}_{min},b^{ij}_{min})$ and $(a^{ij}_{max},b^{ij}_{max})$, which are the positions of diagonally opposite corners of the cell, respectively. 
As depicted in Fig~\ref{fig:matrix}, the geographic area of $A^{T}$ covers the area ahead, behind, and on both sides of the EV and adopts the orientation of the EV. 
% The width and height can be calculated as
% $w=|x^{ij}_{max}-x^{ij}_{min}|$ and $h=|y^{ij}_{max}-y^{ij}_{min}|$.

Each cell $c^{T}_{ij}$ comprises a set of scene objects $O=\{o_1, o_2,..., o_n\}$ that occur within the specific location according to scene representation in the KG. 
Every object $o_k$ has a geo-location, which is represented by its coordinates $(a_k,b_k)$. An object belongs to a cell $c^{T}_{ij}$ when the object coordinates are located within the cell area. We can formally define this as follows:

\begin{equation}
    o_k \in c^{T}_{ij}  \Leftrightarrow  a^{ij}_{min} \leq a_k \leq a^{ij}_{max} \text{ and } b^{ij}_{min} \leq b_k \leq b^{ij}_{max}\text{.}
\end{equation}
As such, the BEV scene representation captures the relative position of the scene objects to the EV and each other in the area matrix. 
% Depending on the focus, the EV can be positioned either in the center row or column of $A_{T}$ (i.e., $c^{T}_{ab}$ where $a=\frac{n}{2}$ and $b=\frac{m}{2}$) or towards the back of the area (e.g., $c^{T}_{ab}$ where $a=\frac{n}{4}$ and $b=\frac{m}{2}$). 
% In our setting, the EV is positioned towards the back of the area (e.g., $c^{T}_{ab}$ where $a=\frac{n}{4}$ and $b=\frac{m}{2}$). 
Every scene object $o_k$ has an object type $e_k$. The object types $E=\{e_1, e_2, ..., e_n\}$ are defined in the ontology and can be referred to by their unique text labels $L=\{l_1, l_2, ..., l_n\}$ respectively.
For the extraction of the scene objects from the KG, we utilize geo-SPARQL queries to fetch the respective objects based on their geo-location. 

\subsection{Serialization}
As Figure~\ref{fig:architecture}b shows, the BEV scene representation, where each cell in the matrix $c^{T}_{ij}$ contains the scene objects $\{o_1, ..., o_n\}$ occurring in the corresponding area, is serialized to a string of tokens.

The serialization method converts two subsequent scenes, $S_T$ and  $S_{T+1}$, into a sequence of tokens. Our approach processes the area matrix $A_T$ in row-major order, starting from $c_{11}^T$ to  $c_{nm}^T$, and then concatenates it with the row-major order of area matrix  $A_{T+1}$, starting from $c_{11}^{T+1}$ to $c_{nm}^{T+1}$.We use the type label $l_k$ of each object $o_k$ occurring in the cell for the encoding.

At the start of the encoding, \textcolor{gray}{<country>}, \textcolor{gray}{<dist>}, and \textcolor{gray}{<orientation\_diff>} tokens are used to indicate that the following token is the country where the EV drives, the traveled distance between scenes $S_{T}$ and $S_{T+1}$, measured in meters, as well as the orientation difference of the EV between the scenes $S_{T}$ and $S_{T+1})$, measured in degrees.
An example is illustrated here, 
\begin{quote}
    ``\textcolor{gray}{<country>} US \textcolor{gray}{<dist>} 4.8 \textcolor{gray}{<orientation\_diff>} 0 \textcolor{gray}{<scene\_start>} lane \textcolor{light-green}{<col\_sep>} lane <concept\_sep> car \textcolor{light-green}{<col\_sep>} pedestrian crossing \textcolor{light-green}{<col\_sep>} walkway ... \textcolor{light-blue}{<row\_sep>} ... \textcolor{light-green}{<col\_sep>} intersection \textcolor{light-green}{<col\_sep>} turn stop area \textcolor{light-green}{<col\_sep>} ...'',
\end{quote}
where \textcolor{light-green}{<col\_sep>} and \textcolor{light-blue}{<row\_sep>} serve as positional delimiters, separating the columns and rows of the matrix in order to preserve the geo-spatial structure of the scene. <concept\_sep> separates different scene objects within a cell.  

\subsection{Scene Learning}

Here, we present two tasks designed to train the FMs for scene comprehension, namely mask prediction and next scene prediction (cf. Figure~\ref{fig:architecture}c). 

\subsubsection{Scene Object Prediction}
The text input formed by a serialized sequence of tokens is randomly masked. An example can be seen as follows,
\begin{quote}
    ``\textcolor{gray}{<country>} US \textcolor{gray}{<dist>} 4.8 \textcolor{gray}{<orientation\_diff>} 0 \textcolor{gray}{<scene\_start>} lane \textcolor{light-green}{<col\_sep>} lane <concept\_sep> car \textcolor{light-green}{<col\_sep>} \textcolor{light-orange}{<$M_1$>} \textcolor{light-green}{<col\_sep>} walkway ... \textcolor{light-blue}{<row\_sep>} ... \textcolor{light-green}{<col\_sep>} \textcolor{light-orange}{<$M_2$>} \textcolor{light-green}{<col\_sep>} turn stop area \textcolor{light-green}{<col\_sep>} ...'',
\end{quote}
where \textcolor{light-orange}{<$M_1$>} and \textcolor{light-orange}{<$M_2$>} are the unique sentinel token used to corrupt the text input. The objective is to predict the dropped-out spans, delimited by the sentinel tokens used to replace them in the input plus a final sentinel token \textcolor{light-orange}{<$M_3$>}, as shown below, 
\begin{quote}
    ``\textcolor{light-orange}{<$M_1$>} pedestrian crossing \textcolor{light-orange}{<$M_2$>} intersection \textcolor{light-orange}{<$M_3$>}''.
\end{quote}

\subsubsection{Next Scene Prediction}
Instead of randomly replacing tokens in the sequence with the sentinel token, all tokens representing map concepts and entities in the next scene $T+1$ are masked. Similar to the previously defined mask prediction task, the learning objective is to predict all these map concepts and entities in the next scene $T+1$, delimited by the sentinel tokens. 

\subsubsection{Learning Loss}
We use cross-entropy loss to maximize the probability of correctly predicting ground truth tokens at their respective positions. Specifically, the loss function is defined as follows:
\begin{equation}
   L = -\sum_{i} y_i \log(\hat{y}_i)\text{,}
\end{equation}
where  $y_i$  is the one-hot encoded ground truth label for $i$-th token while  $\hat{y}_i$  is the predicted probability for $i$-th token.

\section{Experiments}
\subsection{Dataset and Data Augmentation}
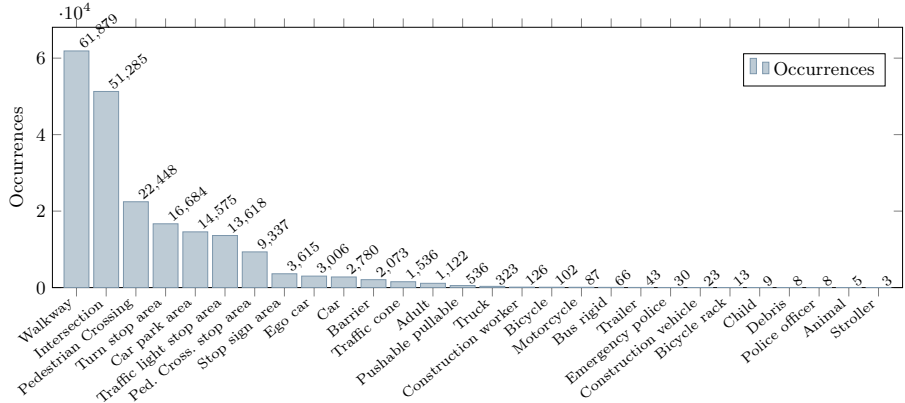
\begin{figure}[tb]
\begin{adjustbox}{width=0.98\textwidth}
\begin{tikzpicture}
\begin{axis}[
    ybar,                      
    bar width=12pt,                 
    enlarge x limits=0.03,         
    symbolic x coords={
        Walkway,
        Intersection,
        Pedestrian Crossing,
        Turn stop area,
        Car park area,
        Traffic light stop area,
        Ped. Cross. stop area,
        Stop sign area,
        Ego car,
        Car,
        Barrier,
        Traffic cone,
        Adult,
        Pushable pullable,
        Truck,
        Construction worker,
        Bicycle,
        Motorcycle,
        Bus rigid,
        Trailer,
        Emergency police,
        Construction vehicle,
        Bicycle rack,
        Child,
        Debris,
        Police officer,
        Animal,
        Stroller,
    },
    xtick=data,                    
    x tick label style={rotate=40, anchor=east, font=\scriptsize}, 
    ymin=0,                        
    xlabel={},             
    ylabel={Occurrences},          
    height=6cm,                    
    width=16cm,                    
    nodes near coords,             
    nodes near coords style={rotate=40, anchor=west, font=\scriptsize},
    % tick label style={font=\scriptsize}, 
    % label style={font=\scriptsize},     
    legend style={at={(0.98,0.9)}, anchor=north east}, % 图例字体缩小
    legend cell align={left},
]

\addplot[draw=light-blue, fill=light-blue!50] coordinates {
        % (Lane,174183)
        (Walkway,61879)
        (Intersection,51285)
        (Pedestrian Crossing,22448)
        (Turn stop area,16684)
        (Car park area,14575)
        (Traffic light stop area,13618)
        (Ped. Cross. stop area,9337)
        (Stop sign area,3615)
        (Ego car,3006)
        (Car,2780)
        (Barrier,2073)
        (Traffic cone,1536)
        (Adult,1122)
        (Pushable pullable,536)
        (Truck,323)
        (Construction worker,126)
        (Bicycle,102)
        (Motorcycle,87)
        (Bus rigid,66)
        (Trailer,43)
        (Emergency police,30)
        (Construction vehicle,23)
        (Bicycle rack,13)
        (Child,9)
        (Debris,8)
        (Police officer,8)
        (Animal,5)
        (Stroller,3)
};

\legend{Occurrences}
\end{axis}
\end{tikzpicture}
\end{adjustbox}
\caption{The occurrence of scene objects in our proposed dataset. The numbers on the bars indicate the occurrence of each corresponding object.}
\label{fig:occurance}
\end{figure}

The BEV symbolic scene representations extracted from the nuScenesKG form the backbone of our training dataset. It encompasses approximately 30,000 driving scenes. To prevent data leakage, we divide these driving scenes into training(80\%), validation(10\%), and testing datasets(10\%). The occurrence of scene objects can be seen in Figure~\ref{fig:occurance}. As the figure illustrates, the majority of scene objects are static objects like walkways, intersections, and pedestrian crossings. 

A random re-mask strategy is employed for data augmentation, i.e., randomly masking tokens within the training data and
reassigning them to different areas (cells) in each scene upon the completion of every epoch. 
This method ensures that the model is continuously exposed to varying prediction scenarios, thereby enhancing its learning and predictive capabilities for diverse driving scenes.

% Given the relatively limited size of this dataset compared to the vast amounts of data typically used to pre-train LLMs, we decided to fine-tune an existing model rather than train one LLM from scratch.

\subsection{Experiment Setting}
\subsubsection{Language Backbone}
We utilize T5, an encoder-decoder language model, to frame language tasks as text-to-text generation problems. 
T5 is available in various sizes: T5-small, T5-base, T5-large, T5-3B, and T5-11B. Our experiments primarily focus on the T5-base model, with additional experiments conducted using T5-small and T5-large to evaluate the impact of model size on performance. Additionally, we include the randomly initialized T5-base model for comparison.

\subsubsection{Scene Range}
In our experiments, the range of the BEV symbolic representation extends 30 m in front of the ego vehicle, 10 m behind it, and 12 m to each side. We intentionally give more focus on the space in front of the car as this has a higher relevance for the actions to be taken. In total, each scene we consider has a height of 40 m and a width of 24 m.

\subsubsection{Number of Areas in Matrix}
The length of the input sequence depends on the resolution of the area matrix (i.e., cell width $w$ and length $l$, and dimensions $n \times m$ of $A$): higher resolutions result in more tokens for the model to process. The pre-trained T5 model can handle sequences of up to 4096 tokens without a significant performance drop. In our setup, we use a matrix dimension of $20 \times 11$, where each cell represents an area of 2 m $\times$ 2 m = 4 m$^2$. 
Additionally, for ablation studies, we use a lower-dimension matrix of 8 $\times$ 5, where each cell corresponds then to an area of 5 m $\times$ 5 m = 25 m$^2$.

\subsubsection{Number of Masked Areas}
The ratio of mask tokens over all tokens during training is important for the learning efficiency of the model. For the masked prediction task, we randomly mask \emph{three} cells in the current scene and \emph{three} cells in the next scene. For the next scene prediction tasks, there is no masking for the current scene, and the entire next scene is masked. Note that we only mask $14 \times 7=98$ central cells. There are two main reasons for that: 1) T5 only supports 100 masking tokens, and increasing the number of predictions would cause a performance drop of T5. 2) It is difficult for the model to predict new incoming objects in the next scene. We replace the margin of the matrix with the empty token <empty> and thus ease the above-mentioned issue. 

\subsubsection{Training Details}
We fine-tune all T5 models using the AdamW optimizer with a learning rate of 0.0001, following a linear schedule without warm up steps. The batch size equals 4. Training for the T5-base model is conducted on a single Nvidia A100 GPU with 80 GB of memory, whereas the T5-large model is trained on a single Nvidia H200 GPU with 141 GB of memory.

% We evaluate the following training approaches to achieve an enhanced scene understanding. 

% \begin{enumerate}
%     \item We use T5 [] as a pretrained LLM and fine-tune it on the scene understanding task. Here we also compare the performance of T5-base and T5-large to evaluate the performance of different model sizes. The models are fine-tuned with 20 epochs with AdamW optimizer of learning rate 0.0001. 
%     \item We use the T5 model architecture, but with random initialization, and train the foundation model on the scene understanding task.
% \end{enumerate}

\subsection{Evaluation Metric}
Accuracy, precision, recall, and F1 are computed to evaluate the performance of our proposed model. 
For the autonomous driving task, it is less risky to predict false positives than to predict false negatives. Predicting a car where none exists may lead to unnecessary avoidance, but failing to predict a car when one is present poses a much higher risk of collision. Hence, recall is a more critical metric for evaluation. All metrics presented in the following sections reflect the performance on the test data.

\subsection{Results}
\subsubsection{Scene Object Prediction}
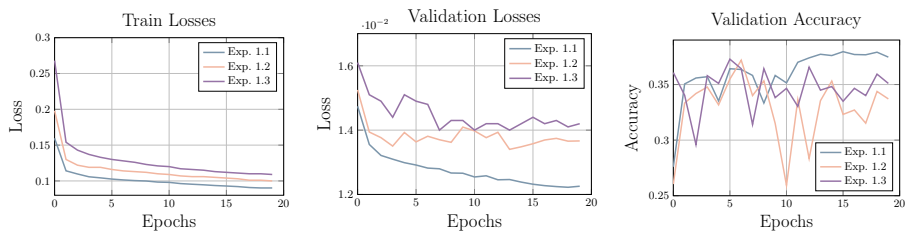
\begin{figure}[tb!]
    \centering
    \begin{subfigure}{0.32\textwidth}
        \centering
            \begin{adjustbox}{width=0.98\textwidth}
            \begin{tikzpicture}
            \begin{axis}[
                title={Train Losses},
                xlabel={Epochs},
                ylabel={Loss},
                title style={font=\Large},
                xlabel style={font=\Large}, % Adjust X-axis label size
                ylabel style={font=\Large}, % Adjust Y-axis label size
                xmin=0, xmax=20,
                ymin=0.08, ymax=0.3,
                legend pos=north east,
                grid=major,
                width=8cm,
                height=6cm,
            ]
            % Plot 1: Line 1
            \addplot[
                very thick,
                color=light-blue
            ]
            coordinates {
            (0,0.1595)
            (1,0.1141)
            (2,0.1097)
            (3,0.1058)
            (4,0.1043)
            (5,0.1026)
            (6,0.1014)
            (7,0.1005)
            (8,0.0999)
            (9,0.0984)
            (10,0.0980)
            (11,0.0963)
            (12,0.0954)
            (13,0.0946)
            (14,0.0936)
            (15,0.0929)
            (16,0.0921)
            (17,0.0908)
            (18,0.0902)
            (19,0.0902)
            };
            
            % Plot 2: Line 2
            \addplot[
                very thick,
                color=light-red
            ]
            coordinates {
            (0,0.198)
            (1,0.130)
            (2,0.122)
            (3,0.119)
            (4,0.119)
            (5,0.116)
            (6,0.114)
            (7,0.113)
            (8,0.112)
            (9,0.110)
            (10,0.109)
            (11,0.107)
            (12,0.106)
            (13,0.106)
            (14,0.105)
            (15,0.104)
            (16,0.103)
            (17,0.101)
            (18,0.101)
            (19,0.100)
            };
            
            % Plot 3: Line 3
            \addplot[
                very thick,
                color=light-purple
            ]
            coordinates {
            (0, 0.268)
            (1, 0.154)
            (2, 0.143)
            (3, 0.137)
            (4, 0.133)
            (5, 0.130)
            (6, 0.128)
            (7, 0.126)
            (8, 0.123)
            (9, 0.121)
            (10, 0.120)
            (11, 0.117)
            (12, 0.116)
            (13, 0.115)
            (14, 0.113)
            (15, 0.112)
            (16, 0.111)
            (17, 0.110)
            (18, 0.110)
            (19, 0.109)
            };
            
            % Legend
            \legend{Exp. 1.1, Exp. 1.2, Exp. 1.3}
            
            \end{axis}
            \end{tikzpicture}
            \end{adjustbox}
        \label{fig:ep_sub1}
    \end{subfigure}
    \hfill
    \begin{subfigure}{0.32\textwidth}
        \centering
            \begin{adjustbox}{width=0.98\textwidth}
            \begin{tikzpicture}
            \begin{axis}[
                title={Validation Losses},
                xlabel={Epochs},
                ylabel={Loss},
                title style={font=\Large},
                xlabel style={font=\Large}, % Adjust X-axis label size
                ylabel style={font=\Large}, % Adjust Y-axis label size
                xmin=0, xmax=20,
                ymin=0.0120, ymax=0.017,
                legend pos=north east,
                grid=major,
                width=8cm,
                height=6cm,
            ]

            % Plot 1: Line 1
            \addplot[
                very thick,
                color=light-blue
            ]
            coordinates {
                (0,0.014741)
                (1,0.013555)
                (2,0.013213)
                (3,0.013097)
                (4,0.012988)
                (5,0.012915)
                (6,0.012820)
                (7,0.012798)
                (8,0.012667)
                (9,0.012660)
                (10,0.012544)
                (11,0.012580)
                (12,0.012456)
                (13,0.012464)
                (14,0.012384)
                (15,0.012318)
                (16,0.012274)
                (17,0.012245)
                (18,0.012223)
                (19,0.012256)
            };
            
            % Plot 2: Line 2
            \addplot[
                very thick,
                color=light-red
            ]
            coordinates {
                (0,0.015251)
                (1,0.013941)
                (2,0.013766)
                (3,0.013504)
                (4,0.013926)
                (5,0.013635)
                (6,0.013810)
                (7,0.013701)
                (8,0.013621)
                (9,0.014094)
                (10,0.013977)
                (11,0.013766)
                (12,0.013933)
                (13,0.013402)
                (14,0.013482)
                (15,0.013577)
                (16,0.013693)
                (17,0.013744)
                (18,0.013657)
                (19,0.013664)
            };
            
            % Plot 3: Line 3
            \addplot[
                very thick,
                color=light-purple
            ]
            coordinates {
                (0,0.0161)
                (1,0.0151)
                (2,0.0149)
                (3,0.0144)
                (4,0.0151)
                (5,0.0149)
                (6,0.0148)
                (7,0.0140)
                (8,0.0143)
                (9,0.0143)
                (10,0.0140)
                (11,0.0142)
                (12,0.0142)
                (13,0.0140)
                (14,0.0142)
                (15,0.0144)
                (16,0.0142)
                (17,0.0143)
                (18,0.0141)
                (19,0.0142)
            };
            
            % Legend
            \legend{Exp. 1.1, Exp. 1.2, Exp. 1.3}
            
            \end{axis}
            \end{tikzpicture}
            \end{adjustbox}
        \label{fig:ep_sub2}
    \end{subfigure}
    \hfill
    \begin{subfigure}{0.32\textwidth}
        \centering
            \begin{adjustbox}{width=0.98\textwidth}
            \begin{tikzpicture}
            \begin{axis}[
                title={Validation Accuracy},
                xlabel={Epochs},
                ylabel={Accuracy},
                title style={font=\Large},
                xlabel style={font=\Large}, % Adjust X-axis label size
                ylabel style={font=\Large}, % Adjust Y-axis label size
                xmin=0, xmax=20,
                ymin=0.25, ymax=0.39,
                legend pos=south east,
                grid=major,
                width=8cm,
                height=6cm,
            ]

            % Plot 1: Line 1
            \addplot[
                very thick,
                color=light-blue
            ]
            coordinates {
                (0,0.276015)
                (1,0.350522)
                (2,0.355861)
                (3,0.357021)
                (4,0.335203)
                (5,0.364217)
                (6,0.363520)
                (7,0.358182)
                (8,0.333346)
                (9,0.358182)
                (10,0.351451)
                (11,0.370019)
                (12,0.373965)
                (13,0.377215)
                (14,0.376054)
                (15,0.379536)
                (16,0.376983)
                (17,0.376750)
                (18,0.379072)
                (19,0.374642)
            };
            
            % Plot 2: Line 2
            \addplot[
                very thick,
                color=light-red
            ]
            coordinates {
                (0,0.260464)
                (1,0.333346)
                (2,0.341934)
                (3,0.347737)
                (4,0.331721)
                (5,0.354932)
                (6,0.371876)
                (7,0.340077)
                (8,0.353308)
                (9,0.315010)
                (10,0.259304)
                (11,0.336364)
                (12,0.283443)
                (13,0.335435)
                (14,0.353075)
                (15,0.323133)
                (16,0.327079)
                (17,0.315242)
                (18,0.343791)
                (19,0.337060)
            };
            
            % Plot 3: Line 3
            \addplot[
                very thick,
                color=light-purple
            ]
            coordinates {
                (0,0.360735)
                (1,0.340309)
                (2,0.295977)
                (3,0.357950)
                (4,0.350986)
                (5,0.372805)
                (6,0.364217)
                (7,0.314081)
                (8,0.364217)
                (9,0.338221)
                (10,0.346576)
                (11,0.330329)
                (12,0.365609)
                (13,0.344952)
                (14,0.348201)
                (15,0.334971)
                (16,0.346576)
                (17,0.340077)
                (18,0.359342)
                (19,0.350986)
            };
            
            % Legend
            \legend{Exp. 1.1, Exp. 1.2, Exp. 1.3}
            
            \end{axis}
            \end{tikzpicture}
            \end{adjustbox}
        \label{fig:ep_sub3}
    \end{subfigure}
    \vspace{-1.5em}
    \caption{\small The losses and accuracy of scene object prediction are reported for Experiment 2, conducted without a dropout strategy, as well as Experiments 3 and 4, which utilized dropout rates of 0.2 and 0.4, respectively. }
    \label{fig:entity prediction}
\end{figure}

To evaluate the spatial understanding ability of  LLMs, we conduct the scene object prediction task. 
As Table~\ref{tab:mask prediction} demonstrates, the baseline that fine-tunes the pre-trained LLM achieves good performance on the object prediction task. 
The overall test accuracy reaches 88.7\%. 
Compared to other LLMs such as LLaMA and ChatGPT, which are widely utilized for scene understanding in autonomous driving, our model demonstrates significantly improved performance.
Experiment 1.1 shows that without the implicit knowledge from a pre-trained LLM, the prediction performance drops significantly from 88.7\% to 37.4\%.  
Figure~\ref{fig:entity prediction} depicts the training/validation loss and validation accuracy. 
Experiment 1.2 and 1.3 show that using drop-out to prevent overfitting does not improve the performance either.

\begin{table}[tb!]
    \centering
    \caption{\small The accuracy, precision, recall, and F1 score for the scene object prediction task. Llama3.1 and ChatGPT are tested in the zero-shot setting. We use the ChatGPT version `2024-02-15-preview'.}
    \scalebox{0.9}{
    \begin{tabular}{l l c c c c}
        \toprule
        \multicolumn{2}{l}{Experiments} & {Accuracy} & Precision & Recall & F1\\ 
        \midrule
        Exp. 1 & - FM4SU(ours)  & 0.887  & 0.866 & 0.744 & 0.786  \\    
        Exp. 2 & - Llama3.1 (8B) & 0.182 & 0.074 & 0.069 & 0.081 \\
        Exp. 2.1 & - Llama3.1 (70B) & 0.224& 0.085 & 0.077 & 0.068\\
        Exp. 3 & - ChatGPT3.5 & 0.353 & 0.181 & 0.123 & 0.141 \\
        Exp. 3.1 & - ChatGPT4o & 0.412 & 0.167 & 0.175 & 0.161 \\
        \midrule
        Exp. 1.1 & - w/o pre-training no dropout & 0.374  & - & - & -      \\    
        Exp. 1.2 & - w/o pre-training dropout rate 0.2 & 0.327 & - & - & -           \\ 
        Exp. 1.3 & - w/o pre-training dropout rate 0.4 & 0.353 & - & - & -          \\ 
        \bottomrule
    \end{tabular}
    }    
    \label{tab:mask prediction}
\end{table}

% Lot of experiments were done during this project. The latest and most meaningful experiments are analysed in this chapter.
% The fine-tuned models were able to reach above 85\% accuracies on both few areas prediction task and next scene prediction task.

\subsubsection{Next Scene Prediction}
We further conduct next scene prediction experiments to evaluate the spatio-temporal understanding ability of pre-trained LLMs in the driving scenarios. 
In our setting, the next scene prediction task aims to predict all scene objects in the next scene instead of just a few. 
Hence, this task is more difficult than the previous tasks. 
As Table~\ref{tab:next scene prediction} demonstrates, the Exp. 4 baseline, which is further fine-tuned in next scene prediction with the checkpoint in Exp. 1, achieves an overall 86.7\% accuracy in the next scene prediction task. 
Compared to the object prediction in Exp. 1, the precision, recall, and F1 score decrease, while the accuracy slightly drops. One potential reason for having high accuracy while low precision, recall, and F1 is that the model predicts the dominant class more frequently, achieving high accuracy but performing poorly on less frequent or minority classes. To further investigate this, we divide the scene objects into two categories, namely dynamic objects and static objects, and evaluate on these two categories. Dynamic objects refer to moving objects on the road, such as cars, motorcycles, and pedestrians, which are relatively rare in the scene. In contrast, static objects are immovable features like lanes, walkways, and pedestrian crossings, which constitute the majority of the scene. As Figure~\ref{fig:dynamic-static-exp1} and \ref{fig:dynamic-static-exp5} shown, the performance of dynamic object prediction in the next scene prediction drops significantly when compared to the scene object prediction task. These results do not surprise, as spatio-temporal relations of dynamic objects are much harder to learn compared to static objects.

\begin{figure}[ht]
    \centering
    \begin{subfigure}{0.32\textwidth}
        \centering
            \begin{adjustbox}{width=0.98\textwidth}
            \begin{tikzpicture}
            \begin{axis}[
                title={Train Losses},
                xlabel={Epochs},
                ylabel={Loss},
                title style={font=\Large},
                xlabel style={font=\Large}, % Adjust X-axis label size
                ylabel style={font=\Large}, % Adjust Y-axis label size
                xmin=0, xmax=10,
                ymin=0.0, ymax=0.5,
                legend pos=north east,
                grid=major,
                width=8cm,
                height=6cm,
            ]
            % Plot 1: Line 1
            \addplot[
                very thick,
                color=light-blue
            ]
            coordinates {
                (1,0.337513)
                (2,0.072988)
                (3,0.051127)
                (4,0.044899)
                (5,0.041571)
                (6,0.038239)
                (7,0.036068)
                (8,0.034476)
                (9,0.033465)
                (10,0.032449)
            };
            
            % Plot 2: Line 2
            \addplot[
                very thick,
                color=light-red
            ]
            coordinates {
                (1,0.473035)
                (2,0.140749)
                (3,0.058655)
                (4,0.047797)
                (5,0.043309)
                (6,0.040557)
                (7,0.037806)
                (8,0.036793)
                (9,0.035200)
                (10,0.034186)
            };
            
            % Plot 3: Line 3
            \addplot[
                very thick,
                color=light-purple
            ]
            coordinates {
                (1,0.470142)
                (2,0.140171)
                (3,0.062131)
                (4,0.051850)
                (5,0.047362)
                (6,0.044031)
                (7,0.041859)
                (8,0.039689)
                (9,0.038095)
                (10,0.038240)
            };
            
            % Legend
            \legend{Exp. 4, Exp. 4.1, Exp. 4.2}
            
            \end{axis}
            \end{tikzpicture}
            \end{adjustbox}
        \label{fig:sub1}
    \end{subfigure}
    \hfill
    \begin{subfigure}{0.32\textwidth}
        \centering
            \begin{adjustbox}{width=0.98\textwidth}
            \begin{tikzpicture}
            \begin{axis}[
                title={Validation Losses},
                xlabel={Epochs},
                ylabel={Loss},
                title style={font=\Large},
                xlabel style={font=\Large}, % Adjust X-axis label size
                ylabel style={font=\Large}, 
                xmin=0, xmax=10,
                ymin=0.005, ymax=0.06,
                legend pos=north east,
                grid=major,
                width=8cm,
                height=6cm,
            ]

            % Plot 1: Line 1
            \addplot[
                very thick,
                color=light-blue
            ]
            coordinates {
                (1,0.020000)
                (2,0.011342)
                (3,0.010067)
                (4,0.009262)
                (5,0.008926)
                (6,0.008591)
                (7,0.008322)
                (8,0.007987)
                (9,0.007852)
                (10,0.007919)
            };
            
            % Plot 2: Line 2
            \addplot[
                very thick,
                color=light-red
            ]
            coordinates {
                (1,0.057450)
                (2,0.013289)
                (3,0.010537)
                (4,0.009664)
                (5,0.009128)
                (6,0.008993)
                (7,0.008591)
                (8,0.008322)
                (9,0.008121)
                (10,0.008054)
            };
            
            % Plot 3: Line 3
            \addplot[
                very thick,
                color=light-purple
            ]
            coordinates {
                (1,0.058523)
                (2,0.013893)
                (3,0.011409)
                (4,0.010537)
                (5,0.010067)
                (6,0.009664)
                (7,0.009195)
                (8,0.009128)
                (9,0.008993)
                (10,0.008792)
            };
            
            % Legend
            \legend{Exp. 4, Exp. 4.1, Exp. 4.2}
            
            \end{axis}
            \end{tikzpicture}
            \end{adjustbox}
        \label{fig:sub2}
    \end{subfigure}
    \hfill
    \begin{subfigure}{0.32\textwidth}
        \centering
            \begin{adjustbox}{width=0.98\textwidth}
            \begin{tikzpicture}
            \begin{axis}[
                title={Validation Accuracy},
                xlabel={Epochs},
                ylabel={Accuracy},
                title style={font=\Large},
                xlabel style={font=\Large}, % Adjust X-axis label size
                ylabel style={font=\Large}, 
                xmin=0, xmax=10,
                ymin=0.38, ymax=0.9,
                legend pos=south east,
                grid=major,
                width=8cm,
                height=6cm,
            ]

            % Plot 1: Line 1
            \addplot[
                very thick,
                color=light-blue
            ]
            coordinates {
                (1,0.751608)
                (2,0.789309)
                (3,0.806048)
                (4,0.818129)
                (5,0.818565)
                (6,0.819585)
                (7,0.831085)
                (8,0.833268)
                (9,0.833705)
                (10,0.835887)
            };
            
            % Plot 2: Line 2
            \addplot[
                very thick,
                color=light-red
            ]
            coordinates {
                (1,0.424388)
                (2,0.762526)
                (3,0.800808)
                (4,0.818130)
                (5,0.809833)
                (6,0.813180)
                (7,0.829339)
                (8,0.832104)
                (9,0.831376)
                (10,0.832977)
            };
            
            % Plot 3: Line 3
            \addplot[
                very thick,
                color=light-purple
            ]
            coordinates {
                (1,0.399934)
                (2,0.704883)
                (3,0.723952)
                (4,0.758742)
                (5,0.771406)
                (6,0.766602)
                (7,0.773443)
                (8,0.774462)
                (9,0.780138)
                (10,0.779992)
            };
            
            % Legend
            \legend{Exp. 4, Exp. 4.1, Exp. 4.2}
            
            \end{axis}
            \end{tikzpicture}
            \end{adjustbox}
        \label{fig:sub3}
    \end{subfigure}
    \caption{\small The training, validation loss and validation accuracy of the next scene prediction task for each Experiment 4, 4.1, 4.2. }
    \label{fig:next scene prediction}
\end{figure}
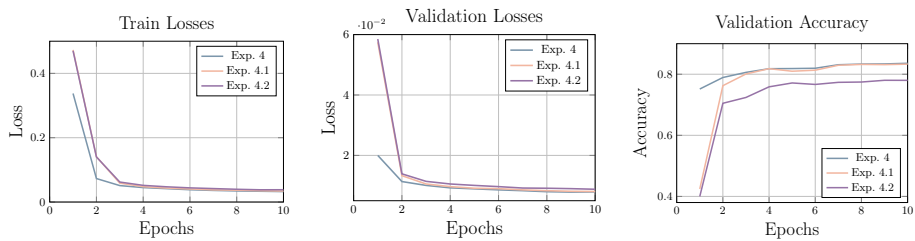

We also investigate if the pre-training of scene object prediction could improve the performance of the next scene prediction. We conduct Exp. 4.1, i.e. fine-tuning on the T5-base checkpoint instead of the checkpoint from Exp. 1.
Comparing Exp. 4 and Exp. 4.1 in Table~\ref{tab:next scene prediction}, the performance does not vary significantly. 
However, as shown in Figure~\ref{fig:next scene prediction}, pre-training on scene object prediction achieves much better performance at the beginning, indicating that spatial prior knowledge from scenes leads to in a faster learning process. 

Additional metadata also plays an important role in temporal scene understanding performance. Experiment 4.2 is conducted by removing all metadata, including country information, as well as the EV’s displacement and orientation shift between consecutive scenes. As Table~\ref{tab:next scene prediction} shows, without these additional metadata, the prediction accuracy, precision, recall, and F1 score drops from 86.7\%, 61.8\%, 59.4\%, 60.3\% to 82.4\%, 59.0\%, 55.1\%, 56.8\%, respectively. The results confirm our intuition, namely that country information is critical due to differences in driving orientation (right-hand drive vs. left-hand drive), and the ego vehicle’s displacement and orientation shift are essential for understanding the relative movement of the surrounding environment. 

\begin{table}[tb!]
\centering
\caption{\small The Accuracy, Training loss, Precision, Recall, and F1 Score for the Task of Next Scene Prediction.}
\scalebox{0.9}{
\begin{tabular}{l l c c c c}
\toprule
\multicolumn{2}{l}{Experiments} & Accuracy & Precision & Recall & F1 \\ 
\midrule
Exp. 4 &- FM4SU (Next Scene Prediction) & 0.867  & 0.618 & 0.594 & 0.603 \\ 
Exp. 4.1 &- w/o Scene Object Prediction Training &  0.865 & 0.622 & 0.598 & 0.608 \\ 
Exp. 4.2 &- w/o Additional Metadata & 0.824 & 0.590 & 0.551 & 0.568\\ 
\bottomrule
\end{tabular}
}
\label{tab:next scene prediction}
\end{table}

\begin{figure}[tb!]
    \centering
    \begin{subfigure}[b]{0.49\textwidth}
        \centering
            \begin{adjustbox}{width=0.98\textwidth}
            \begin{tikzpicture}
            \begin{axis}[
                ybar,
                bar width=25pt,
                width=9cm,
                height=6cm,
                enlarge x limits=0.3,
                legend style={at={(1,0)}, anchor=south east, legend columns=1},
                legend cell align={left},
                ylabel={Accuracy},
                xlabel={Element Type},
                title style={font=\large},
                xlabel style={font=\large}, % Adjust X-axis label size
                ylabel style={font=\large}, % Adjust Y-axis label size
                symbolic x coords={Dynamic, Static},
                xtick=data,
                ymin=0, ymax=1.0,
                nodes near coords,
                nodes near coords align={vertical}
            ]
            \addplot [draw=light-blue, fill=light-blue!50] coordinates {(Dynamic, 0.84) (Static, 0.89)};
            \addplot [draw=light-red, fill=light-red!50] coordinates {(Dynamic, 0.63) (Static, 0.86)};
            \addplot [draw=light-purple, fill=light-purple!50] coordinates {(Dynamic, 0.70) (Static, 0.87)};
            
            \legend{Precision, Recall, F1}
            \end{axis}
            \end{tikzpicture}
            \end{adjustbox}
        \caption{\small Experiment 1}
        \label{fig:dynamic-static-exp1}
    \end{subfigure}
    %\hfill
    \begin{subfigure}[b]{0.49\textwidth}
        \centering
            \begin{adjustbox}{width=0.98\textwidth}
            \begin{tikzpicture}
            \begin{axis}[
                ybar,
                bar width=25pt,
                width=9cm,
                height=6cm,
                enlarge x limits=0.3,
                legend style={at={(1,0)}, anchor=south east, legend columns=1},
                legend cell align={left},
                ylabel={Accuracy},
                xlabel={Element Type},
                title style={font=\large},
                xlabel style={font=\large}, % Adjust X-axis label size
                ylabel style={font=\large}, % Adjust Y-axis label size
                symbolic x coords={Dynamic, Static},
                xtick=data,
                ymin=0, ymax=1.0,
                nodes near coords,
                nodes near coords align={vertical}
            ]
            \addplot [draw=light-blue, fill=light-blue!50] coordinates {(Dynamic, 0.42) (Static, 0.81)};
            \addplot [draw=light-red, fill=light-red!50] coordinates {(Dynamic, 0.39) (Static, 0.80)};
            \addplot [draw=light-purple, fill=light-purple!50] coordinates {(Dynamic, 0.40) (Static, 0.81)};
            
            \legend{Precision, Recall, F1}
            \end{axis}
            \end{tikzpicture}
            \end{adjustbox}
        \caption{Experiment 4}
        \label{fig:dynamic-static-exp5}
    \end{subfigure}
    \vfill
    \begin{subfigure}[b]{0.49\textwidth}
        \centering
            \begin{adjustbox}{width=0.98\textwidth}
            \begin{tikzpicture}
            \begin{axis}[
                ybar,
                bar width=25pt,
                width=9cm,
                height=6cm,
                enlarge x limits=0.3,
                legend style={at={(1,0)}, anchor=south east, legend columns=1},
                legend cell align={left},
                ylabel={Accuracy},
                xlabel={Element Type},
                title style={font=\large},
                xlabel style={font=\large}, % Adjust X-axis label size
                ylabel style={font=\large}, % Adjust Y-axis label size
                symbolic x coords={Dynamic, Static},
                xtick=data,
                ymin=0, ymax=1.0,
                nodes near coords,
                nodes near coords align={vertical}
            ]
            \addplot [draw=light-blue, fill=light-blue!50] coordinates {(Dynamic, 0.43) (Static, 0.82)};
            \addplot [draw=light-red, fill=light-red!50] coordinates {(Dynamic, 0.39) (Static, 0.80)};
            \addplot [draw=light-purple, fill=light-purple!50] coordinates {(Dynamic, 0.41) (Static, 0.81)};
            
            \legend{Precision, Recall, F1}
            \end{axis}
            \end{tikzpicture}
            \end{adjustbox}
        \caption{\small Experiment 4.1}
        \label{fig:dynamic-static-exp6}
    \end{subfigure}
     %\hfill
    \begin{subfigure}[b]{0.49\textwidth}
        \centering
            \begin{adjustbox}{width=0.98\textwidth}
            \begin{tikzpicture}
            \begin{axis}[
                ybar,
                bar width=25pt,
                width=9cm,
                height=6cm,
                enlarge x limits=0.3,
                legend style={at={(1,0)}, anchor=south east, legend columns=1},
                legend cell align={left},
                ylabel={Accuracy},
                xlabel={Element Type},
                title style={font=\large},
                xlabel style={font=\large}, % Adjust X-axis label size
                ylabel style={font=\large}, % Adjust Y-axis label size
                symbolic x coords={Dynamic, Static},
                xtick=data,
                ymin=0, ymax=1.0,
                nodes near coords,
                nodes near coords align={vertical}
            ]
            \addplot [draw=light-blue, fill=light-blue!50] coordinates {(Dynamic, 0.40) (Static, 0.78)};
            \addplot [draw=light-red, fill=light-red!50] coordinates {(Dynamic, 0.34) (Static, 0.76)};
            \addplot [draw=light-purple, fill=light-purple!50] coordinates {(Dynamic, 0.36) (Static, 0.77)};
            
            \legend{Precision, Recall, F1}
            \end{axis}
            \end{tikzpicture}
            \end{adjustbox}
        \caption{Experiment 4.2}
        \label{fig:dynamic-static-exp7}
    \end{subfigure}
    \caption{\small The precision, recall, and F1 score for dynamic/static objects of the scene object prediction (Exp. 1) as well as the next scene prediction task (Exp. 4.1, 4.2, and 4.3). The number of dunamic and static object are 9374 and 371271, respectively.}
  
    \label{fig:experiment1}
\end{figure}
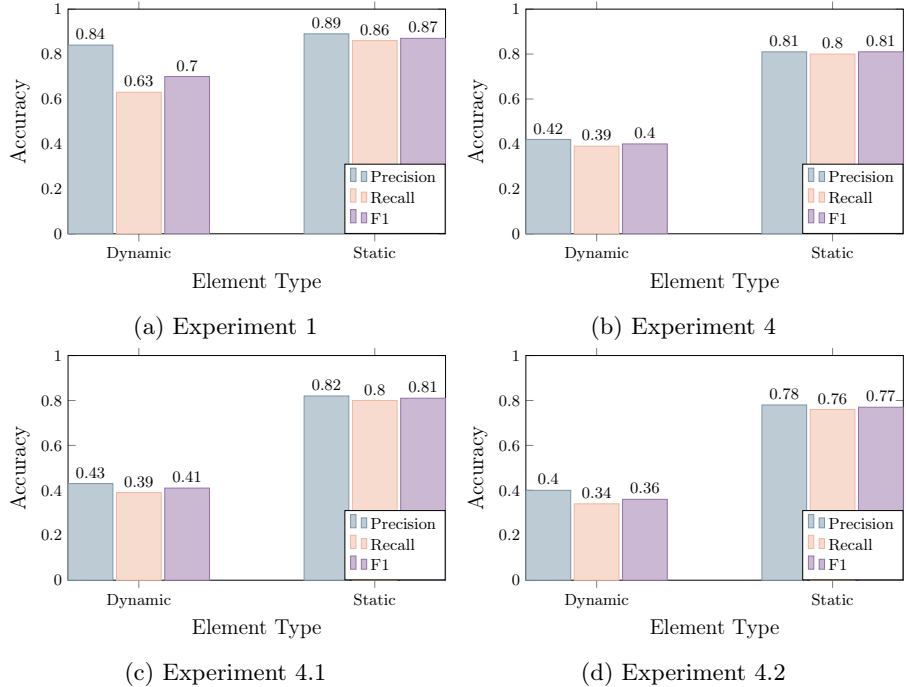

\subsection{Ablation Study}
\subsubsection{Scene Resolution}
We conduct experiments with different scene resolutions for scene object detection tasks. 
Table~\ref{tab:scene resolution} demonstrates the baseline with scene resolution $20 \times 11$ performs significantly better compared to resolution $8 \times 5$. 
This shows that the higher granularity of information helps the LLM to reach a higher prediction accuracy. Due to their lower resolution, larger cells encompass more scene objects and thus increase the complexity of the prediction task.
\begin{table}[htb!]
    \centering
    \caption{\small The scene object prediction accuracy with different scene resolution. Experiment 1 and 1.4 are trained with a scene resolution of $20 \times 11$ and $8 \times 5$, respectively.
    }
    \scalebox{0.9}{
    \begin{tabular}{l l c}
        \toprule
        \multicolumn{2}{l}{Experiments} & {Accuracy}  \\ 
        \midrule
        Exp. 1 &- Baseline (Scene Resolution $20 \times 11$) & 0.887 \\     
        Exp. 1.4 & - Scene Resolution $8 \times 5$ & 0.396 \\    
        % Exp 3. w/o pre-training dropout rate 0.2 & 0.327           & 0.0137 & - & - & -           \\ 
        % Exp 4. w/o pre-training dropout rate 0.4 & 0.353           & 0.0141  & - & - & -          \\ 
        \bottomrule
    
    \end{tabular}
    }
    \label{tab:scene resolution}
\end{table}
\subsubsection{Model Size}
\begin{table}[h!]
    \centering
    \caption{\small The next scene prediction accuracy with different model sizes. 
    Experiment 4.3, 4, 4.4 are trained with T5-small, T5-base, T5-large language backbones, respectively.
    }
    \scalebox{0.9}{
    \begin{tabular}{l l c}
        \toprule
        
        \multicolumn{2}{l}{Experiments} & {Accuracy} \\ 
        \midrule
        Exp. 4.3 & - with T5-small Backbone & 0.770 \\ 
        Exp. 4 & - Baseline (with T5-base Backbone) & 0.867 \\     
        Exp. 4.4 &- with T5-large Backbone & 0.861\\    
        % Exp 3. w/o pre-training dropout rate 0.2 & 0.327           & 0.0137 & - & - & -           \\ 
        % Exp 4. w/o pre-training dropout rate 0.4 & 0.353           & 0.0141  & - & - & -          \\ 
        \bottomrule
    \end{tabular}
    }
    \label{tab:model size}
\end{table}
To examine whether model size influences the performance of the scene prediction task, we conducted experiments using language backbones of varying sizes, specifically T5-small, T5-base, and T5-large. 
Table~\ref{tab:model size} shows larger models generally demonstrate better performance. 
However, given the marginal difference in performance between T5-large and T5-base, we primarily use T5-base for most experiments, as it is more resource-efficient in terms of GPU usage.

\section{Conclusion and Future Work}
This paper introduced FM4SU, a novel approach for training a FM for scene understanding in autonomous driving.
The objective is to overcome the limitations of current methods in comprehending the complex the spatio-temporal evolution of scenery information. 
The methodology utilizes KGs to capture sensory observations and domain knowledge, as well as a novel BEV symbolic representation of each scene extracted from KG. 
This BEV scene representation includes the spatio-temporal information among the objects across the scenes. 
Next, this representation is serialized into a sequence of tokens and given to pre-trained language models for learning an inherent understanding of the co-occurrence among scene elements and generating predictions on the next scenes.
FM4SU is evaluated using the T5 model and demonstrates superior performance in both scene object prediction and next scene prediction task, with an accuracy of 88.7\% and 86.7\%, respectively. 
% Moreover, the ablation studies highlight that our approach is indeed leveraging the knowledge from the pre-trained model to solve the autonomous driving tasks since the pre-trained T5 performed significantly better compared to the non-pre-trained version.
Based on our results, we conclude that FM4SU provides a promising framework for developing larger and more exhaustive FMs for scene understanding, and opens avenues for future exploration and downstream tasks in the context of autonomous driving.
Further, as a part of our contribution, we have released the BEV scene representation dataset for the nuScenesKG as well as the source code for the extraction and training in order to invite the research community to develop and benchmark new algorithms and FMs for scene understanding based on the nuScenes dataset.

For future work, we strive to integrate into FM4SU other LLMs and perform a wide range of investigations on the performance and scalability aspects. 
Further, our aim is to assess its performance on other downstream tasks, for example, 3D object detection and trajectory prediction with by incorporating the acquired scene understanding.

\bibliographystyle{splncs04}
\bibliography{ref}





% LNCS Homepage, \url{http://www.springer.com/lncs}, last accessed 2023/10/25
% \end{thebibliography}
\end{document}